\def\year{2015}
\documentclass[letterpaper]{article}
\usepackage{aaai}
\usepackage{times}

\usepackage{url}
\usepackage{amsmath}
\usepackage{amssymb}
\newtheorem{example}{Example}
\newtheorem{definition}{Definition}
\newtheorem{theorem}{Theorem}
\newtheorem{lemma}{Lemma}

\usepackage{color}
\usepackage{comment}
\usepackage{bm}

\usepackage{helvet}
\usepackage{courier}
\begin{document}
%
\title{Towards Log-Linear Logics with Concrete Domains}
\author{Melisachew Wudage Chekol \and Jakob Huber \and Heiner Stuckenschmidt \\
Data and Web Science Group\\
Universit{\"a}t Mannheim \\
Mannheim, Germany \\
\{mel,jakob,heiner\}@informatik.uni-mannheim.de}
\maketitle
\begin{abstract}
\begin{quote}
We present $\mathcal{MEL}^{++}$ (M denotes Markov logic networks) an extension of the log-linear description logics $\mathcal{EL}^{++}$-LL with  concrete domains, nominals, and instances. We use Markov logic networks (MLNs) in order to find the most probable, classified and coherent $\mathcal{EL}^{++}$ ontology from an $\mathcal{MEL}^{++}$ knowledge base. In particular, we develop a novel way to deal with concrete domains (also known as datatypes) by extending MLN's cutting plane inference (CPI) algorithm.  
\end{quote}
\end{abstract}

\section{Introduction}
In description logics (DLs) a concrete domain is a construct that can be used to define new classes by specifying restrictions on attributes that have literal values (as opposed to relationships to other concepts). 
Practical applications of DLs usually require concrete properties with values from a fixed domain, such as strings or integers, supporting built-in predicates. For DLs that are extended with concrete domains, there exist partial functions mapping objects of the abstract domain to values of the concrete domain, and can be used for building complex concepts.
Concrete domains can be used to construct complex concepts as for instance, 
the axiom $Teenager \equiv Person \sqcap \exists age. (\geq, 13) \sqcap \exists age. (\leq,19)$ defines a teenager as a person whose age is at least 13 and at most 19. In DLs, concrete domains are also known as \textit{datatypes}. Several probabilistic extensions of DLs opt to exclude datatypes while, in fact, it is an essential feature as several knowledge extraction tools produce weighted rules or axioms that contain concrete data values. Reasoning over these data either to infer new knowledge or to verify correctness is indispensable. Additionally, recent advances in information extraction have paved the way for the automatic construction and growth of large,  semantic knowledge bases from different sources.  However, the very nature of these extraction techniques entails that the resulting knowledge bases
may  contain  a  significant  amount  of  incorrect,  incomplete,  or  even
inconsistent (i.e.,  uncertain) knowledge,  which makes efficient  reasoning and query  answering  over  this  kind  of  uncertain   data  a challenge.  To address these issues, there exist ongoing studies on probabilistic knowledge bases.

The study of extending DLs to handle uncertainty and vagueness has gained momentum recently. There have been several proposals to add probabilities to various DLs. Probabilistic DLs can be classified in several dimensions. One possible classification is on the reasoning mechanism used: Markov logic networks (MLNs), Bayesian networks, and probabilistic reasoning. There exist some studies that employ MLNs to extend various DLs.  
The study in \cite{lukasiewicz-et-al:2012} extends $\mathcal{EL}^{++}$ with probabilistic uncertainty based on the annotation of axioms using MLNs. The main focus of this work is ranking queries in	descending order of probability of atomic inferences which is different from the objective of this paper.
Another study in
\cite{niepert-et-al:2011},  presents a probabilistic extension of the DL $\mathcal{EL}^{++}$ without nominals and concrete domains in MLN in order to find the most probable coherent ontology. In doing so, they have developed a reasoner for probabilistic OWL-EL called ELOG \cite{noessner-et-al:2011}. In this study, we extend this work in order to deal with concrete domains in addition to nominals and instances. In databases, MLNs have been used to create a probabilistic datalog called  Datalog$+/-$.
It is an extension of datalog that allows to express ontological axioms by using rule-based constraints \cite{gottlob-et-al:2013}. 
The probabilistic extension of Datalog$+/-$ uses MLNs as the underlying probabilistic semantics. The focus of this work is on scalable threshold query answering which is different from that of this work.

Other literatures extend DLs with Bayesian networks. Some notable works include: an extension of $\mathcal{EL}$ with Bayesian networks called $\mathcal{BEL}$ is presented in \cite{ceylan-et-al:2014}. They study the complexity of reasoning under $\mathcal{BEL}$ to show that reasoning is intractable. However, their work does not discuss probabilities in the ABox and concrete domains are excluded. On the other hand, 
in \cite{damato-et-al:2008}, they added  uncertainty to DL-Lite based on Bayesian networks. Additionally, they have shown that satisfiability test and query answering in probabilistic DL-Lite can be reduced to satisfiability test and query answering in the DL-Lite family.	Further, it is proved that satisfiability checking and union of conjunctive query answering can be done in LogSpace in the data complexity.

Consequently, as discussed above, most of the studies that involve extending description logics to deal with uncertainty  by using either Bayesian or MLNs often excluded concrete domains. This is partly due to either the lack of supporting features or the difficulty in dealing with them.
In this paper, we study a novel way of dealing with uncertainty involving concrete domains. Henceforth, we provide an extension to $\mathcal{EL}^{++}$-LL  with concrete domains, nominals and instances.

\section{Preliminaries}
In this section, we present a brief summary of: $\mathcal{EL}^{++}$, Markov logic networks, cutting plane inference, and $\mathcal{EL}^{++}$-LL. For a detailed discussion on these subjects, we refer the reader to \cite{baader-et-al:2005,richardson-et-al:2006,riedel:2012,niepert-et-al:2011} and the references therein. 
\subsection{$\bm{\mathcal{EL}^{++}}$}
$\mathcal{EL}^{++}$ is the description logic underlying the OWL 2 profile OWL-EL\footnote{\url{http://www.w3.org/TR/owl2-profiles/}}. 
\subsubsection{Syntax}
Given a set of concept names $\mathrm{N_C}$, role names $\mathrm{N_R}$, individuals $\mathrm{N_I}$, and feature names $\mathrm{N_F}$,
$\mathcal{EL}^{++}$ concepts and roles are formed according to the following syntax:
\begin{align*}
C &~::= \top \mid \bot \mid A \mid C \sqcap D \mid \exists R.C \mid \{a\} \mid \exists F.r
\end{align*}
A concept in $\mathcal{EL}^{++}$ is either a top, bottom concept, an atomic concept or a complex concept (formed by conjunction and existential restriction).
Given a datatype restriction $r=(o,v)$ and $x\in \mathcal{D}$, we say that $x$ satisfies $r$ and  write $r(x)$ iff $(x,v) \in o$, where $o 
\in \{<,\leq,>,\geq,=\}$, $o$ is interpreted as the standard relation on real numbers, and $\mathcal{D} \subseteq \mathbb{R}$ is a concrete domain \cite{despoina-et-al:2011}. In this work, we consider only numerical concrete domains and leave out the others for future work. 
An $\mathcal{EL}^{++}$ TBox contains a set of GCI (General Concept Inclusion) axioms, i.e., $C \sqsubseteq D$, as well as role inclusion axioms, i.e., $R_1 \circ \cdots \circ R_k \sqsubseteq R$. 

The semantics of $\mathcal{EL}^{++}$ concepts and roles is given by an interpretation function $\mathcal{I}=(\Delta^\mathcal{I},.^{\mathcal{I}})$ which consists of a non-empty (abstract) domain $\Delta^\mathcal{I}$ and a mapping function $.^\mathcal{I}$ \cite{baader-et-al:2005}. 

\subsubsection{Semantics}
The semantics of $\mathcal{EL}^{++}$ concepts and roles is given by an interpretation function $\mathcal{I}=(\Delta^\mathcal{I},.^{\mathcal{I}})$ which consists of a non-empty (abstract) domain $\Delta^\mathcal{I}$ and a mapping $.^\mathcal{I}$ that assigns to each atomic concept $A \in \mathrm{N_C}$ a subset of $\Delta^\mathcal{I}$, to each abstract role $R \in \mathrm{N_R}$ a subset of $\Delta^\mathcal{I} \times \Delta^\mathcal{I}$, to each concrete relation $F \in \mathrm{N_F}$ a subset of $\Delta^\mathcal{I} \times \mathcal{D}$, and to each individual $a \in \mathrm{N_I}$ an element of $\Delta^\mathcal{I}$. The mapping $\cdot^\mathcal{I}$ is extended to all concepts and roles as follows: 
\begin{align*}
(\top)^\mathcal{I} &~= \Delta^\mathcal{I} \\
(\bot)^\mathcal{I} &~= \emptyset \\ 
(\{a\})^\mathcal{I} &~= \{a^\mathcal{I}\} \\
(C \sqcap D )^\mathcal{I} &~= C^\mathcal{I} \cap D^\mathcal{I} \\
(\exists R.C) &~= \{x \in \Delta^\mathcal{I} \mid \exists y\in \Delta^\mathcal{I}:\\&~~~~~~~~(x,y)\in R^\mathcal{I} \wedge y\in C^\mathcal{I}\}\\
(\exists F.r)^\mathcal{I} &~= \{x\in \Delta^\mathcal{I} \mid \exists v\in \mathcal{D}: (x,v) \in F^\mathcal{I} \\
&~~~~~~~~\wedge r(v)\} \\
(C \sqsubseteq D)^\mathcal{I} &~= C^\mathcal{I} \subseteq D^\mathcal{I} \\
(R_1 \circ \cdots \circ R_k \sqsubseteq R)^\mathcal{I} &~= R_1^\mathcal{I} \circ \cdots \circ R_k^\mathcal{I} \subseteq R^\mathcal{I}
\end{align*}
%
Knowledge about specific objects can be expressed using concept and role
assertions of the form $C(a)$ and $R(a,b)$. The axioms and assertions are contained in
the TBox and ABox, respectively, which together form a knowledge base (KB).
An $\mathcal{EL}^{++}$  knowledge base (or ontology) $\mathcal{O}=(\mathcal{T},\mathcal{A})$ consists of a set $\mathcal{T}$ of general concept inclusion
axioms (TBox) and role inclusion axioms, and possibly a set
$\mathcal{A}$ of assertional axioms (ABox). A concept name
$C$ in an ontology $\mathcal{O}$, is \emph{unsatisfiable} iff, for each interpretation
$\mathcal{I}$ of $\mathcal{O}$, $C^\mathcal{I}=\emptyset$. An ontology $\mathcal{O}$
is \emph{incoherent} iff there exists an unsatisfiable concept name $C$ in $\mathcal{O}$, i.e., $C \models \bot$ \cite{flouris-et-al:2006}.

To simplify the translation of probabilistic $\mathcal{EL}^{++}$ KB into FOL, we first obtain the \textit{normal} form of the KB in such a way that satisfiability is preserved \cite{baader-et-al:2005,krotzsch:2011}.
An $\mathcal{EL}^{++}$ KB is in \textit{normal} form if its axioms are in the following form:
\begin{alignat*} {4}
&~ C(a) \quad && R(a,b) \quad && A \sqsubseteq \bot  \quad &&   \top \sqsubseteq C\\
&~ A \sqsubseteq \{c\} \quad && \{a\} \sqsubseteq \{c\} \quad && A \sqsubseteq C \quad && A \sqcap B \sqsubseteq C \\
&~ \exists R.A \sqsubseteq C \quad && A \sqsubseteq \exists R.B \quad && A \sqsubseteq \exists F.r \quad && \exists F.r \sqsubseteq A \\
&~ R_1 \sqsubseteq R_2 \quad && R_1 \circ R_2 \sqsubseteq R
\end{alignat*}
where $A,B,C \in \mathrm{N_C}, R,R_1,R_2 \in \mathrm{N_R}, F \in \mathrm{N_F}$, $r$ is a datatype restriction, and $a,b,c \in \mathrm{N_I}$.

It is possible to provide a probabilistic extension of $\mathcal{EL}^{++}$ using MLNs.
An $\mathcal{EL}^{++}$ KB can be seen as a set of hard constraints on the set of possible interpretations: 
if an interpretation violates even one axiom or assertion, it has zero probability. The basic idea in MLNs is to 
soften these constraints, i.e., when an interpretation violates one axiom or assertion in the KB it is less probable, but 
not impossible. The fewer axioms an interpretation violates, the more probable it becomes. Each axiom and assertion
has an associated weight that reflects how strong a constraint is: the higher the weight, the greater 
the difference in log probability between an interpretation that satisfies the axiom and one that does not, other 
things being equal \cite{richardson-et-al:2006}. 
\subsection{Markov Logic Networks}
Markov Logic Networks (MLNs) combine Markov networks
and  first-order  logic (FOL)  by  attaching  weights  to  first-order  formulas  and
viewing  these  as  templates  for  features  of  Markov  networks \cite{richardson-et-al:2006}.
 An MNL $L$ is a set of pairs $(F_i,w_i)$ where $F_i$ is a formula in FOL and $w_i$ is a real number representing a weight. Together with a finite set of constants $C$, it defines a Markov Network $M_{L,C}$, where $M_{L,C}$ contains one node for each possible grounding of each predicate appearing in $L$. The value of the node is $1$ if the ground predicate is true, and $0$ otherwise. The probability distribution over possible worlds $x$ specified by the ground Markov network $M_{L,C}$ is given by: 
$$P(X=x) = \dfrac{1}{Z} \mathrm{exp}\big(\sum\limits_{i=1}^F w_i n_i(x)\big)$$
where $F$ is the number of formulas in the MLN and $n_i(x)$ is the number of true groundings of $F_i$ in $x$.  The groundings of a formula are formed simply by replacing its variables with constants in all
possible ways. 
The \textit{Herbrand Universe} $H$ for an MLN $L$ is the set of all
terms that can be constructed from the constants in $L$. The \textit{Herbrand Base} $\mathrm{HB}$ is often defined as the set of all ground predicates (atoms) that can be constructed using the predicates in $L$ and the terms in $H$.
In this paper we focus on MLNs whose formulas are function-free clauses.

In order to compute a maximum a-posteriori (MAP) state of an MLN, we
formulate the problem as an integer linear program (ILP) using the cutting plane inference algorithm.

\subsection{Cutting Plane Inference (CPI)}
A MAP query corresponds to an optimization problem with linear constraints and a linear objective function. Hence, it can be formulated and solved as an instance of an integer linear program (ILP). \cite{riedel:2012,noessner-et-al:2013}~introduced cutting plane inference as a meta algorithm that transforms an MLN into ILP. The basic idea of CPI is to add all constraints to the ILP that violate the current intermediate solution. This process is repeated until no (additional) violated ground clauses exist. An ILP solver resolves the conflicts by computing an optimal truth assignment for an MLN. Hence, the solution of the final ILP corresponds to the MAP state. It is necessary to execute several iterations as the intermediate solution changes after each iteration and more violated clauses might be detected. At the beginning of each CPI iteration it is necessary to determine the violated ground clauses $\mathcal{G}$ that are specified by the MLN and are in conflict with the intermediate solution. A binary ILP variable $x_{\ell} \in \{0,1\}$ gets assigned to each grounded predicate occurring in a violated clause $g \in \mathcal{G}$. The value of the the variable $x_{\ell}$ is $1$ if the respective literal $\ell$ is true and $0$ when it is false. These variables are used to generate ILP constraints that are added to the ILP for each violated ground clause. For each clause $g \in \mathcal{G}$, we define ${L}^{+}(g)$ as the set of ground atoms that occur unnegated in $g$ and ${L}^{-}(g)$ as the set of ground atoms that occur negated in $g$. The transformation scheme depends on the weight $w_g \in \mathbb{R}$  of the violated clause $g$. It is also necessary to create a binary variable $z_g$ for every $g$ with $w_g \neq \infty$ that is used in the objective of the ILP. For every ground clause $g$ with $w_g > 0$, the following constraint has to be added to the ILP.
\begin{equation*}
    \displaystyle\sum_{\ell \in {L}^{+}(g)} x_{\ell} + \displaystyle\sum_{\ell \in {L}^{-}(g)} (1-x_{\ell}) \geq z_g
\end{equation*}
A ground atom $\ell$ that is set to false (true if it appears negated) by evidence will not be included in the ILP as it cannot fulfil the respective constraint. For every $g$ with weight $w_g <0$, we add the following constraint to the ILP:
\begin{equation*}
    \displaystyle\sum_{\ell \in {L}^{+}(g)} x_{\ell} + \displaystyle\sum_{\ell \in {L}^{-}(g)} (1-x_{\ell}) \leq (|{L}^{+}(g)|+|{L}^{-}(g)|)  z_g
\end{equation*}
The variable $z_g$ expresses if a ground formula $g$ is true considering the optimal solution of the ILP. However, for every $g$ with weight $w_g = \infty$ this variable can be replaced with 1 as the respective formula cannot be violated in any solution:
\begin{equation*}
    \displaystyle\sum_{\ell \in L^{+}(g)} x_{\ell} + \displaystyle\sum_{\ell \in L^{-}(g)} (1-x_{\ell})    \geq 1
\end{equation*}

Finally, the objective of the ILP sums up the weights of the (satisfied) ground formulas:
\begin{equation*}
    \max \displaystyle\sum_{g \in \mathcal{G}} w_g z_g
\end{equation*}
The MAP state corresponds to the solution of the ILP in the last CPI iteration. It can be directly obtained from the solution as the assignment of the variables $x_\ell$ can be directly mapped to the optimal truth values for the ground predicates, i.e., $x_i = \texttt{true}$ if the corresponding ILP variable is $1$ and $x_i = \texttt{false}$ otherwise.
The MAP state of an $\mathcal{EL}^{++}$-LL TBox can be computed by a reduction into CPI.

\subsection{$\bm{\mathcal{EL}^{++}}$-LL}
 $\mathcal{EL}^{++}$-LL (Log-linear $\mathcal{EL}^{++}$) is a probabilistic extension of $\mathcal{EL}^{++}$ without nominals, instances and concrete domains \cite{niepert-et-al:2011}. Each $\mathcal{EL}^{++}$-LL TBox axiom is either deterministic (i.e., axioms that are known to be true) or uncertain (i.e., axioms that have a degree of confidence). The uncertain axioms have associated weight. Formally, a $\mathcal{EL}^{++}$-LL TBox is given by $\mathcal{T}=(\mathcal{T}^D, \mathcal{T}^U)$, where $\mathcal{T}^D$ and $\mathcal{T}^U$, is a set of pairs of $\langle S,w_S\rangle$ where $S$ is an axiom and $w_S$ is its real-valued weight, denote deterministic and uncertain axioms respectively.

The semantics of an $\mathcal{EL}^{++}$-LL TBox is given by a joint probability distribution over a \emph{coherent} $\mathcal{EL}^{++}$ TBox. Given TBoxes $\mathcal{T}=(\mathcal{T}^D, \mathcal{T}^U)$ and $\mathcal{T}'$ over the same vocabulary, the probability of $\mathcal{T}'$ is given by:
\begin{align*}
P(\mathcal{T}') &~ = \begin{cases}
\dfrac{1}{Z} \mathrm{exp}\bigg(\sum\limits_{\{\forall(S,w_S)\in \mathcal{T}^U:\mathcal{T}'\models S\}} w_S \bigg) \\ ~~~~~~~\text{if }~ \mathcal{T}' \models \mathcal{T}^D \wedge \mathcal{T}' \not\models \bot \\
0 ~~~~~\text{otherwise}
\end{cases}
\end{align*}

In order to generate the most probable, coherent and classified TBox using MLN, $\mathcal{EL}^{++}$ completion rules and $\mathcal{EL}^{++}$-LL TBox axioms are translated into FOL formulae.
In the following, we show how to extend $\mathcal{EL}^{++}$-LL with nominals, instances, and concrete domains.

\section{Extending $\bm{\mathcal{EL}^{++}}$-LL with Nominals, Instances and Concrete Domains}
In \cite{niepert-et-al:2011}, the authors claim that their approach is extensible to the Horn fragments of DLs (look \cite{krotzsch:2011} for instance). 
To take advantage of this claim, we extend $\mathcal{EL}^{++}$-LL with probabilistic knowledge expressed through nominals, individuals, and concrete domains. The syntax of this extension (that we call $\mathcal{MEL}^{\mathrm{++}}$) is the  same as that of $\mathcal{EL}^{++}$-LL, basically, 
 it is the syntax of $\mathcal{EL}^{++}$ with weights attached to each uncertain axiom and assertion. An $\mathcal{MEL}^{++}$ KB has two components: deterministic $\mathrm{KB}^D$ and uncertain $\mathrm{KB}^U$ knowledge bases. 
 In order to provide semantics, we assume that $\mathrm{KB}^D$ is coherent.   The semantics of \emph{coherent} $\mathcal{MEL}^{\mathrm{++}}$ KBs is given by a probability distribution as defined below. 
  \begin{definition}
 Given an $\mathcal{MEL}^{\mathrm{++}}$ knowledge base $\mathrm{KB}=(\mathrm{KB}^D, \mathrm{KB}^U)$ over a vocabulary of $\mathrm{N_C}$, $\mathrm{N_R}$, $\mathrm{N_F}$, and $\mathrm{N_I}$, the semantics of a \emph{coherent} $\mathrm{KB}_i=(\mathrm{KB}^D_i, \mathrm{KB}^U_i)$ over the same vocabulary is given by a probability distribution:
 \begin{align*}
 P(\mathrm{KB}') &~= \begin{cases}
\dfrac{1}{Z}\mathrm{exp}\bigg(\sum\limits_{\{\forall (o_j,w_j)\in \mathrm{KB}^U:\mathrm{KB}_i\models o_j\}} w_j\bigg) \\ ~~~~~~~~~\text{if}~~  \mathrm{KB}_i \models \mathrm{KB}^D \wedge \mathrm{KB}_i \not\models \bot \\
 0 ~~~~~~\text{otherwise}
 \end{cases}
 \end{align*}
 \end{definition}
 \begin{example} Consider an $\mathcal{MEL}^{++}$ $\mathrm{KB} = (\mathrm{KB}^D, \mathrm{KB}^U)$:
\begin{align*}
 \mathrm{KB}^{D} =&~ \{~ Toddler \sqcap Adult \sqsubseteq \bot\}, \\
 \mathrm{KB}^{U} = &~\{\langle Toddler \sqsubseteq ~\exists age.(\leq, 3), ~0.8 \rangle, \\
  &~~~~\langle \exists age.(\leq, 3) \sqsubseteq Person, ~0.7 \rangle, \\
  &~~~~\langle Toddler \sqsubseteq Adult, ~0.1 \rangle, ~\langle age(john, 2), ~0.7\rangle \} 
 \end{align*}
 The probabilities of the axioms and assertions can be computed as follows:
 \begin{align*}
 P\big(\{Toddler \sqsubseteq ~\exists age.(\leq, 3)\}\big) =&~  \dfrac{1}{Z}\mathrm{exp}(0.8) \\
  P\big(\{Toddler \sqsubseteq Adult\}\big)= &~  0 \\
  P\bigg(\{Toddler \sqsubseteq ~\exists age.(\leq, 3), age(john, 2), &~ \\ ~~~~~~\exists age.(\leq, 3) \sqsubseteq Person\}\bigg) =&~  \dfrac{1}{Z}\mathrm{exp}(2.2) \\
  P\big(\{\}\big) =&~ \dfrac{1}{Z}\mathrm{exp}(0) \\
  P\big(\{Toddler \sqcap Adult \sqsubseteq \bot\}\big)= &~  1 \\
  Z =~ \mathrm{exp}(0.8) + \mathrm{exp}(2.2) + \mathrm{exp}(0.7) + \mathrm{exp}(0)
 \end{align*}
 \end{example}
 In order to derive the most probable, classified and coherent $\mathcal{EL}^{++}$ ontology from an $\mathcal{MEL}^{++}$ KB, we transform the KB, TBox completions rules \cite{baader-et-al:2005}, concrete domains, and ABox completion rules \cite{krotzsch:2011} into FOL formulae. 
\subsection{Nominals}
(Un)certain axioms that contain nominals can be translated into FOL in MNL by using Definition \ref{def:mapping}. Inference in MNL can be done by 
converting the completion rule CR6 \cite{baader-et-al:2005} into FOL 
and enforcing that each nominal $a_i \in \mathrm{N_I}$ is distinct. Alternatively, 
\textit{unique name assumption} for individuals names can be enforced by using the axiom 	$\{a\} \sqcap \{b\} \sqsubseteq \bot$ for all relevant individual names $a$ and $b$.
 In addition, the transformation of TBox completion rules into FOL in MNL is given in Table \ref{tab:tboxCompletionRules}.

By using nominals, instance knowledge can be added to an ABox.

\subsection{ABox}
Since the description logic $\mathcal{EL}^{++}$ is equipped with nominals. ABox knowledge can be converted into TBox axioms. 	
Thus, with nominals, ABox becomes syntactic sugar:
$$C(a) \Leftrightarrow \{a\} \sqsubseteq C,~~R(a,b)	 \Leftrightarrow \{a\} \sqsubseteq \exists R.\{b\}$$ Instance checking in turn is directly reducible to subsumption checking in the presence of nominals.
There exist two ways to represent uncertain ABox assertions, i.e., $C(a)$ and $R(a,b)$, in MLN: 
\begin{itemize}
\item[i.] transform ABox assertions into TBox axioms using nominals as follows: 
\begin{align*}
\langle C(a), w_1\rangle \Leftrightarrow &~ \langle \{a\} \sqsubseteq C, w_1\rangle \\
\langle R(a,b), w_2\rangle \Leftrightarrow&~ \langle \{a\} \sqsubseteq \exists R.\{b\}, w_2\rangle 
\end{align*}
\item[iii.] introduce two new predicates for each instance type as:
\begin{align*}
\langle C(a), w_1\rangle &~ \mapsto inst(a,C) ~~~w_1 \\
\langle R(a,b), w_2\rangle &~ \mapsto rinst(a,R,b) ~~~w_2
\end{align*}
This approach requires transforming ABox completion rules into FOL, so as to generate classified ontologies. 
\end{itemize}
In this paper, we consider the second approach (ii)\footnote{We leave a comparison of the two approaches  as a future work.}. Next, we show how concrete domains are translated into the MLN framework. 
\begin{table*}[t!]
\centering
\renewcommand{\arraystretch}{1.4}
\begin{tabular}{|ll|}
\hline
$F_1$ -- $F_9$ & Refer to Table 2 in \cite{niepert-et-al:2011}. \\
$F_{10}$& $\forall c,d,a,r: \mathrm{subNom}(c,a) \wedge \mathrm{subNom}(d,a) \wedge \mathrm{rsup}(c,r,d) \rightarrow \mathrm{sub}(c,d)$ \\ 
$F_{11}$& $\forall c,d,a,r,b: \mathrm{subNom}(c,a) \wedge \mathrm{subNom}(d,a) \wedge \mathrm{rsupNom}(b,r,d) \rightarrow \mathrm{sub}(c,d)$ \\
$F_{12}$& $\forall c,d,f,o,v: \mathrm{sub}(c,d) \wedge \mathrm{rsupEx}(d,f,o,v) \Rightarrow \mathrm{rsupEx}(c,f,o,v)$ \\ 
$F_{13}$& $\forall c,d,f,o,v: \mathrm{rsupEx}(c,f,o_1,v_1) \wedge \mathrm{rsubEx}(f,o_2,v_2,d) \wedge \mathrm{eval}(o_1,v_1,o_2,v_2) \Rightarrow \mathrm{sub}(c,d)$ \\ \hline
\end{tabular}
\caption{TBox completion rules.}
\label{tab:tboxCompletionRules}
\end{table*}

\begin{table*}[t!]
\centering
\renewcommand{\arraystretch}{1.4}
\begin{tabular}{|ll|}
\hline
$F_{14}$ & $\forall x,A,B: \mathrm{inst}(x,A) \wedge \mathrm{sub}(A,B) \Rightarrow \mathrm{inst}(x,B)$ \\ 
$F_{15}$ & $\forall x,A_1,A_2,B: \mathrm{inst}(x,A_1) \wedge \mathrm{inst}(x,A_2) \wedge \mathrm{int}(A_1,A_2,B) \Rightarrow \mathrm{inst}(x,B)$ \\  
$F_{16}$ & $\forall x,y,R,A,B: \mathrm{rinst}(x,R,y) \wedge \mathrm{inst}(y,A) \wedge \mathrm{rsub}(A,R,B) \Rightarrow \mathrm{inst}(x,B)$ \\ 
$F_{17}$ & $\forall x,y,R,S: \mathrm{rinst}(x,R,y) \wedge \mathrm{psub}(R,S) \Rightarrow \mathrm{rinst}(x,R,y)$ \\ 
$F_{18}$ & $\forall x,y,z,R_1,R_2,R_3: \mathrm{rinst}(x,R_1,y) \wedge \mathrm{rinst}(y,R_2,z) \wedge  \mathrm{pcomp}(R_1,R_2,R_3) \Rightarrow \mathrm{rinst}(x,R_3,z)$ \\ 
$F_{19}$ & $\forall x,a,B: \mathrm{ninst}(x,a) \wedge \mathrm{inst}(x,B) \Rightarrow \mathrm{inst}(a,B)$ \\ 
$F_{20}$ & $\forall x,a,B: \mathrm{ninst}(x,a) \wedge \mathrm{inst}(a,B) \Rightarrow \mathrm{inst}(x,B)$ \\ 
$F_{21}$ & $\forall x,a,z,R: \mathrm{ninst}(x,a) \wedge \mathrm{rinst}(z,R,x) \Rightarrow \mathrm{rinst}(z,R,a)$ \\ 
$F_{22}$ & $\forall x,A,B: \mathrm{sub}(\top,A) \wedge \mathrm{inst}(x,B) \Rightarrow \mathrm{inst}(x,A)$ \\ 
$F_{23}$ & $\forall x,x',R,A,B: \mathrm{inst}(x,a) \wedge \mathrm{rsup}(A,R,B) \Rightarrow \mathrm{rinst}(x,R,x')$ \\ 
$F_{24}$ & $\forall x,x',R,A,B: \mathrm{inst}(x,a) \wedge \mathrm{rsup}(A,R,B) \Rightarrow \mathrm{inst}(x',B)$ \\ 
$F_{25}$& $\forall f,op,v,C: \mathrm{rsupEx}(f,op,v,C) \wedge \mathrm{rinst}(a,f,v') \wedge \mathrm{eval}(v,op,v') \Rightarrow \mathrm{inst}(a,A)$ \\ 
$F_{26}$& $\forall a,A, f,v: \mathrm{inst}(a,A) \wedge \mathrm{rsubEx}(A,f,=,v) \Rightarrow \mathrm{rinst}(a,f,v)$ \\ 
$F_{27}$& $\forall a, A_1, A_2, f, v: \mathrm{inst}(a,A_1) \wedge \mathrm{inst}(a,A_2) \wedge \mathrm{intEx}(A_1,A_2,f,op,v) \Rightarrow \mathrm{rinst}(a,f,v)$ \\ \hline
\end{tabular}
\caption{ABox completion rules.}
\label{tab:aboxCompletionRules}
\end{table*}

\subsection{Concrete Domains	}

Reasoning over uncertain concrete domains can be done by transforming the datatype predicates in the axioms and assertions into  mixed integer programming as shown in \cite{straccia:2012}. However, in this work, we introduce an efficient approach that transforms the predicates into a test function that evaluates to \textit{true} or \textit{false} based on the grounding generated by an extension of the CPI algorithm.  
Inference involving axioms that contain concrete domains can be done according to the  deduction rules given below: 
\begin{align*}
&~\frac{A \sqsubseteq B ~~~~ B \sqsubseteq \exists F.(o,v)}{A \sqsubseteq \exists F.(o,v)} \\
&~ \frac{A \sqsubseteq \exists F.(o_1,v_1) ~~~\exists F.(o_2,v_2) \sqsubseteq B}{A \sqsubseteq B}  ~~~ eval(o_1,v_1,o_2,v_2)\\
&~\frac{\exists F.(o,v_1) \sqsubseteq A ~~~F(a,v_2)}{A(a)}  ~~~ eval(o,v_1,=,v_2) \\
&~\frac{A(a) ~~~A \sqsubseteq \exists F.(=,v)}{F(a,v)} 
\end{align*}
where $eval(\ldots)$ checks if all possible values of the first \textit{operator-value} pair $(o_1,v_1)$ are covered by the possible values of the second \textit{operator-value} pair $(o_2,v_2)$, when so, it evaluates to true otherwise false. The  function $eval(\ldots)$ is defined based on a datatype $\mathcal{D}$, i.e., $\mathbb{N}$ or $\mathbb{Z}$ or $\mathbb{R}$, and algebraic operators. Some of the algebraic comparisons, computed via $eval(\ldots)$, that are useful to determine inference are listed below: 
\begin{align*}
eval(\leq,v_1,<,v_2) &~ :=  v_1 < v_2 \\
eval(\leq,v_1,\leq,v_2) &~ :=  v_1 \leq v_2 \\
eval(=,v_1,<,v_2) &~ :=  v_1 < v_2 \\
eval(=,v_1,\leq,v_2) &~ :=  v_1 \leq v_2 \\
eval(=,v_1,=,v_2) &~ :=  v_1 = v_2 \\
eval(=,v_1,\geq,v_2) &~ :=  v_1 \geq v_2 \\
eval(=,v_1,>,v_2) &~ :=  v_1 > v_2 \\
eval(\geq,v_1,\geq,v_2) &~ :=  v_1 \geq v_2 \\
eval(\geq,v_1,>,v_2) &~ :=  v_1 > v_2 \\
eval(>,v_1,>,v_2) &~ :=  v_1 \geq v_2 
\end{align*}
This function is computed on-demand after each CPI iteration as discussed in the next section.
The translation of the deduction rules into FOL is given in Table \ref{tab:tboxCompletionRules} and Table~\ref{tab:aboxCompletionRules}.
\begin{example}\label{ex:InferenceDatatype}
Consider an $\mathcal{MEL}^{++}$ $\mathrm{KB}=\{\langle 2YearOld \sqsubseteq  \exists age.(=,2), 0.7\rangle, \langle \exists age.(\leq,3) \sqsubseteq Toddler, 0.8\rangle\}$ that contains axioms expressed using concrete domains. 
From the KB, the axiom $2YearOld \sqsubseteq Toddler$ can be inferred since $eval(o_1,v_1,o_2,v_2)$ is \textit{true}, i.e.,  $eval(=, 2, \leq,3) := 2\leq 3$. 
\end{example}
So far we have discussed how axioms and assertions can be translated into FOL. Next, we show how the most probable KB is derived using MAP inference.

\section{Computing a Most Probable KB}
To derive the most probable classified and coherent ontology from a weighted $\mathcal{EL}^{++}$ KB, we proceed by transforming TBox and ABox completion rules, schema axioms, and assertions into function-free FOL formulae. The formulae corresponding to the translation of completion rules into FOL are shown in Table \ref{tab:tboxCompletionRules} and Table \ref{tab:aboxCompletionRules}. The formulae from $F_1$ through $F_9$ are taken from \cite{niepert-et-al:2011}. Additionally,
a \textit{bijective}  mapping function is provided in Definition \ref{def:mapping} to transform axioms and assertions into formulae. 
Of particular interest for us is proposing a novel way to deal with concrete domains under MLN by modifying the Cutting Plane Inference (CPI) algorithm.

In $\mathcal{EL}^{++}$,  it is possible to build incoherent TBox axioms due to the presence of the bottom concept $\bot$, for instance, consider the axiom $\{a\} \sqsubseteq \bot$, this cannot be satisfied by any interpretation. To filter out such incoherencies in models generated by MLN, we include the formula $\forall c: \neg sub(c,\bot)$ (formula $F_9$ in Table \ref{tab:tboxCompletionRules}) to the translation of the completion rules into FOL. This technique has already been used in \cite{niepert-et-al:2011}.  

\begin{definition}\label{def:mapping}[Mapping $\mathcal{MEL}^{++}$ KB into Ground FOL predicates]
The function $\varphi$ translates a normalized $\mathcal{MEL}^{++}$ knowledge base KB into FOL formulae in MLN as follows:
\allowdisplaybreaks
\begin{align*}
 C(a) \mapsto &~ \mathrm{inst}(a,C)   \\
 R(a,b) \mapsto &~ \mathrm{rinst}(a,R,b) \\
 A \sqsubseteq \bot \mapsto &~ \mathrm{sub}(A,\bot) \\
 \top \sqsubseteq C \mapsto &~  \mathrm{sub}(\top,C) \\
 A \sqsubseteq \{c\} \mapsto &~ \mathrm{subNom}(A,\{c\}) \\
 \{a\} \sqsubseteq \{c\} \mapsto &~ \mathrm{sub}(\{a\},\{c\}) \\
 A \sqsubseteq C \mapsto &~ \mathrm{sub}(A,C) \\
 A \sqcap B \sqsubseteq C \mapsto &~ \mathrm{int}(A,B,C) \\ 
 \exists R.A \sqsubseteq C \mapsto &~ \mathrm{rsub}(A,R,C) \\
A\sqsubseteq \exists R.B \mapsto &~  \mathrm{rsup}(A,R,B) \\
 A\sqsubseteq \exists F.(o,v) \mapsto &~ \mathrm{rsupEx}(A,F,o,v) \\
 \exists F.(o,v) \sqsubseteq A \mapsto &~ \mathrm{rsubEx}(F,o,v,A) \\
 R_1 \sqsubseteq R_2 \mapsto &~ \mathrm{psub}(R_1,R_2)\\
 R_1 \circ R_2 \sqsubseteq R \mapsto &~ \mathrm{pcom}(R_1,R_2,R) \\
 int(\{a_i\}, \{a_j\}, \bot) &~~~\text{where }a_i, a_j \in \mathrm{N_I} \text{ and } i \not= j
\end{align*}
where $a,b,c \in \mathrm{N_I}$, $A,B,C \in \mathrm{N_C}$, $R, R_1, R_2 \in\mathrm{N_R}$, $F \in \mathrm{N_F}$, $o \in \{<,\leq,>,\geq,=\}$, and $v \in \mathbb{R}$ (set of real numbers).
\end{definition}
\begin{lemma}
The translation of an $\mathcal{EL}^{++}$ KB  into FOL and vice versa can be done in polynomial time in the size of the knowledge base \cite{lukasiewicz-et-al:2012}.
\end{lemma}
From the above Lemma, we see that the translation of $\mathcal{MEL}^{++}$ KB completion rules, axioms, and assertions into FOL in MLN does not affect the complexity of inference in MLN. Besides, as
\textit{typed variables} and \textit{constants} greatly reduce size of ground
Markov nets. We introduce types to all of the predicates shown in Tables \ref{tab:tboxCompletionRules} and Table \ref{tab:aboxCompletionRules}.

\begin{theorem}
Given an $\mathcal{MEL}^{++}$ ontology $\mathrm{KB}=(\mathcal{T},\mathcal{A})$ and $\mathrm{KB}' \subseteq \mathrm{KB}$, a Herbrand interpretation $\mathcal{H}$ is a model of $\mathrm{KB}'$, i.e., $\mathcal{H} \models \mathrm{KB}'$ if and only if there exist a mapping function $\varphi$ such that $\varphi(\mathcal{H}) \models \mathrm{KB}'$. 
\end{theorem}

So far we have introduced a mapping function $\varphi$ for KB assertions and axioms and  completion rules as formulae ($F_1$--$F_{27}$). The next step requires using MAP inference of MLN to obtain the most probable ontology of a given $\mathcal{MEL}^{++}$ KB.

\subsection{Maximum A-Posteriori Inference (MAP)}
In order to deal with $\mathcal{MEL}^{++}$ datatypes, we introduced a predicate called $eval(\ldots)$ in the translation of $\mathcal{EL}^{++}$ completion rules into FOL, depicted in Table \ref{tab:tboxCompletionRules} and Table \ref{tab:aboxCompletionRules}. The truth value of $eval(\ldots)$ is computed by evaluating the logical expressions corresponding to datatypes in an $\mathcal{MEL}^{++}$ KB. For instance, consider the $eval(\ldots)$ predicate in Example \ref{ex:InferenceDatatype}. In the following, we show how the expression $(=, 2) \subseteq (\leq, 3)$, operator-value pair coverage, i.e., is evaluated by extending the CPI algorithm. 
%
Thus, we propose an extension of CPI by incorporating algebraic expressions. In particular, our extension addresses a limitation of MLN with respect to concrete domains.
 In general, all (numerical) values are represented as constants in MLN. The only semantics that are related to constants might be the type
to which they belong. This enables more efficient grounding and leads to smaller
MLNs. However, this does hardly cover the characteristics of numerical values. Therefore, we exploit the iterative character of CPI in order to evaluate numerical (in)equalities. The extension can be considered
as additional features that are only used on-demand. It is formula-specific as it
affects the ground values and the truth value of specific constraints. Hence, it can
be implemented as an extension of the detection of the violated constraints.

The algorithm identifies at the beginning of each CPI iteration for each formula all
violated groundings considering the current intermediate solution. Each of the violated
ground clauses has to be translated and added to the ILP. Therefore, an ILP variable is
generated for each ground predicate as well as additional ILP constraints. Datatype ground predicates $eval(\ldots)$ 
 appear during this process as any other predicates. However, we exploit
there semantics to decide whether $eval(\ldots)$ predicates evaluate to
\textit{true} or \textit{false}. Depending on the result of the evaluation of the attached boolean expression of the respective predicate, we decide whether it is necessary to add the violated
ground clause to the ILP. For instance, if the datatype predicate is positive
(negative) and it appears without negation  (or negation) in the formula, we do not add the ground
clause to the ILP as it is not violated in the current iteration. Otherwise, we need
to add the clause to the ILP but leave out the datatype ground predicates as they can
not fulfil the violated clause, i.e., the respective literal is false due to evidence. Hence,
we do not introduce ILP variables for datatype predicates as they will not be added to
the ILP.  Instead, we compute the truth value of the datatype predicates on-the-fly and
only on-demand. Hence, the proposed approach improves the efficiency of processing
numerical predicates in a Markov logic solver without sacrificing the performance.
We implemented this algorithm as an extension to the MLN inference engine ROCKIT\footnote{\url{https://code.google.com/p/rockit/}} \cite{noessner-et-al:2013}. We leave out testing this implementation with different ontologies as a future work.

\begin{theorem}\label{thm:soundness}
Given the following: 
\begin{itemize}
\item an $\mathcal{MEL}^{++}$ knowledge base $\mathrm{KB} = (\mathrm{KB}^D, \mathrm{KB}^U)$ formed from a vocabulary containing a finite set of individuals $\mathrm{N_I}$, concepts $\mathrm{N_C}$, features $\mathrm{N_F}$, and roles $\mathrm{N_R}$,
\item $\mathrm{HB}$ as a Herbrand base of the formulae $F$ in Table \ref{tab:tboxCompletionRules} and Table \ref{tab:aboxCompletionRules} over the same vocabulary,
\item $G_1$ as a set of ground formulae constructed from  $\mathrm{KB}^D$, and
\item  $G_2$ as a set of ground formulae constructed from $\mathrm{KB}^U$,
\end{itemize}
the most probable coherent and classified ontology is obtained with: 
$$\varphi^{-1}(\hat{I}) = \underset{\mathrm{HB} \supseteq I \models G_1 \cup F}{\arg\max}\bigg(\sum_{(o_j,w_j) \in G_2:I \models o_j} w_j \bigg) $$		
\end{theorem}

From Theorem \ref{thm:soundness} and the results in \cite{roth:1996}, finding the most probable, classified and coherent $\mathcal{MEL}^{++}$ KB is in NP. The \textit{hardness} of this complexity bound can be obtained by reducing partial weighted MaxSAT  problem 
into an $\mathcal{MEL}^{++}$ MAP query. 
Consequently, the MAP problem for $\mathcal{MEL}^{++}$ is NP-hard. 

\section{Conclusion}
In this work, we have extended $\mathcal{EL}^{++}$-LL into $\mathcal{MEL}^{++}$ with nominals, concrete domains and instances. 
In particular, we proposed an extension to the CPI algorithm in order to deal with reasoning under uncertain concrete domains. We have implemented the proposed approach and  planned to carry out experiments in the future. We will also investigate to extend the proposed approach to other datatypes  such as Date, Time, and so on.
\bibliographystyle{aaai}

\end{document}